\documentclass[letterpaper, 10 pt, conference]{ieeeconf}  
\IEEEoverridecommandlockouts
\pdfoutput=1
\usepackage[T1]{fontenc}
\usepackage[english]{babel}
\usepackage{graphicx}   
\usepackage{epsfig}             
\usepackage{amsmath}            
\usepackage{amssymb}            
\usepackage{mathtools}
\usepackage{multicol}
\usepackage[bookmarks=true]{hyperref}
\usepackage{xcolor}
\usepackage{siunitx}
\usepackage{lipsum}
\usepackage{scrextend}
\usepackage{algorithm}
\usepackage{algpseudocode}
\usepackage{multirow}
\usepackage{cite}
\usepackage[inkscapelatex=false]{svg}

\usepackage{bm}

\usepackage{MnSymbol}
\usepackage{mathdots}
\usepackage{amsthm}
\usepackage{empheq}

\theoremstyle{plain}

\theoremstyle{definition}

\theoremstyle{remark}

\newcommand{\R}{\mathbb{R}}

\newcommand{\Rn}[1]{\R^{#1}}

\newcommand{\Rnm}[2]{\R^{#1 \times #2}}

\usepackage{siunitx}

\usepackage[caption=false,font=footnotesize]{subfig}

\usepackage{ctable}

\title{
\bfseries{Deployment-Ready UWB Localization for Industrial Ground Robots with Automatic Anchor Calibration and Terrain-Aware Fusion}
}

\author{Alexander Raab$^{\dagger}$, Giulio Delama$^{\ddagger}$, Roland Jung$^{\ddagger}$ and Stephan Weiss$^{\ddagger}$
\thanks{*This work was supported by the FFG project LUMEIK-5G (923407), funded under the "GigaApp, 2. Call" program by the Austrian Federal Ministry of Innovation, Mobility and Infrastructure (BMIMI).}%
\thanks{*This work was supported by the Austrian Ministry of Climate Action and Energy (BMK), grant agreement 905337 (ROBComm).}%
\thanks{$^{\dagger}$Alexander Raab is with AGILOX Services GmbH, Austria {\tt\small alexander.raab@agilox.net}}%
\thanks{$^{\ddagger}$Giulio Delama, Roland Jung and Stephan Weiss are with the Control
	of Networked Systems Group, University of Klagenfurt, Austria 
        {\tt\small \{firstname.lastname\}@ieee.org}}%
}

\hypersetup{
  pdfinfo={
    Title={},
    Author={},
    Subject={},
    Keywords={}
  }
}

\begin{document}
\bstctlcite{BSTcontrol}

\maketitle
\thispagestyle{empty}
\pagestyle{empty}


\begin{abstract}
Ultra-Wideband (UWB) ranging has become a viable option for industrial Autonomous Mobile Robot (AMR) localization due to improved accuracy and low cost.
However, real-world deployments remain limited by two recurring challenges: calibrating static anchors can be time-consuming and error-prone, and integrating UWB with existing onboard sensors requires careful design to ensure robust and consistent pose estimation.
Addressing these challenges, this paper presents an end-to-end pipeline that combines automatic anchor calibration with a generic multi-sensor estimator tailored to surface-bound vehicle motion. It targets existing AMR stacks in scenarios where robot pose priors are available for initialization. The calibration stage estimates anchor positions and range biases, while the localization stage fuses UWB with proprioceptive sensing in a bias-aware Extended Kalman Filter to improve consistency without extensive parameter tuning.
Experiments on a commercial logistics AMR in a warehouse setting demonstrate accurate positioning indoors and across outdoor transitions, with improved consistency compared to an earlier estimator formulation. Evaluation on an independent forklift dataset further indicates transferability to other platforms. The method remains effective in test cases with limited line-of-sight and sparse anchor coverage.
These results show that UWB localization can be deployed with substantially reduced manual effort while preserving the accuracy required for industrial AMRs.
The collected warehouse dataset is made publicly available.\footnote{The dataset is available at \textit{http://sst.aau.at/cns/datasets}. \newline \hspace{-.3em}\textbf{Preprint version, accepted May/2026 (CASE), DOI to follow ©IEEE.}}
\end{abstract}
\section{Introduction}
\label{sec:intro}

Due to the growing demand for industrial automation, Autonomous Mobile Robots (AMRs) have become central to modern logistics and manufacturing \cite{Fragapane2022AMRs}.
Accurate and robust localization is essential for these robots, enabling efficient, fully autonomous navigation in dynamic environments and across indoor-outdoor transitions.
Although laser- and vision-based localization methods remain common, they are comparatively expensive and power-intensive. Ultra-Wideband (UWB) is therefore increasingly used as a cost-effective and reliable alternative, particularly in Global Navigation Satellite System (GNSS)-denied facilities such as warehouses and factories \cite{Elsanhoury2022LocReview}.
However, two common challenges limit their deployment in real-world facilities.
First, anchor calibration often requires substantial manual effort, auxiliary systems, or carefully planned trajectories. Second, integrating UWB with onboard sensors requires explicit handling of measurement interdependencies, sensor noise, and systematic biases to maintain accurate and consistent estimates. This typically entails extensive parameter tuning, especially when artificial state constraints are used. In addition, the limited availability of public datasets for UWB-based AMR localization restricts benchmarking and method development.

\begin{figure}[]
    \centering
    \includegraphics[width=\linewidth, trim={0cm 2cm 0cm 0},clip]
    {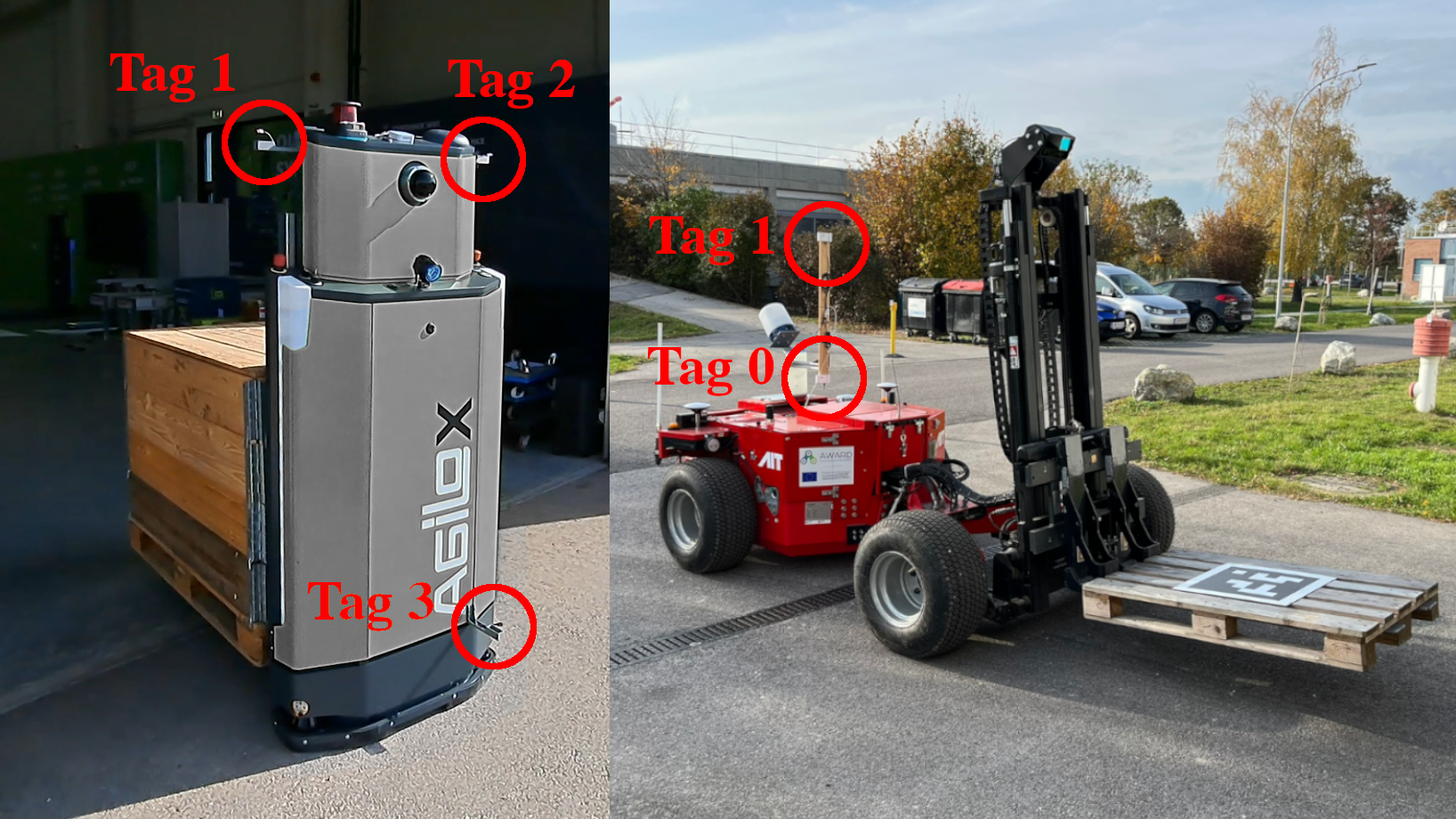}
    \includegraphics[width=\linewidth, trim={3.5cm 11.5cm 3.5cm 12cm},clip]{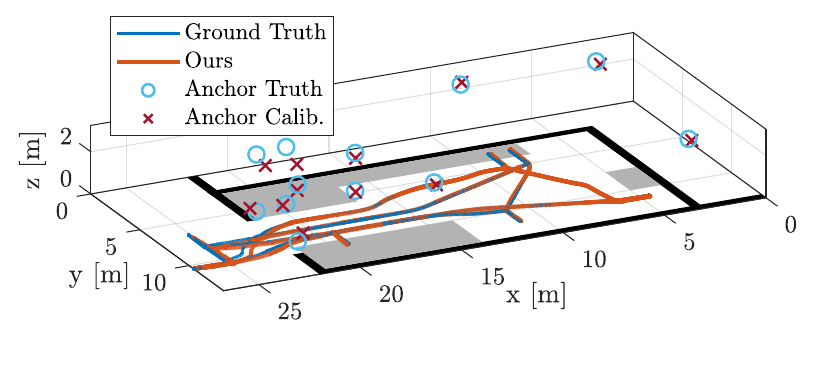}
    \vspace*{-12mm}
    \caption{Depiction of the mobile robots utilized for real-world evaluation. The proposed calibration and fusion pipeline was deployed with the logistics AMR (left), while the generalization of the discussed estimator adaptations are evaluated based on data of the autonomous forklift (right). The trajectory below visualizes the central components of our strategy, combining auto-calibration for anchor positions with UWB-odometry fusion for localization.}
    \label{fig:robots}
    \vspace{-1.5em}
\end{figure}

To address these limitations, this paper proposes an end-to-end UWB positioning pipeline that integrates a real-time UWB anchor initialization approach~\cite{Delama2023UVIO:Initialization,Jung2024ModularCalibrationNew} with a terrain-informed Extended Kalman Filter (EKF) framework for ground vehicles \cite{Raab2025MESEKF}. The pipeline enables AMRs to localize in previously unvisited areas, including indoor–outdoor transitions where reliance on indoor positioning alone is insufficient.
Automatic initialization of UWB anchors is performed in a single step using robot poses supplied by the AMR and UWB ranges collected during an initial calibration trajectory.
To improve localization, the original UWB measurement formulation is extended with sensor states that capture calibration uncertainty as well as constant and range-dependent biases.
Real-world tests on a commercial logistics AMR, shown in Fig.~\ref{fig:robots}, demonstrate single-trajectory anchor calibration and accurate localization without manual adaptation of UWB calibration or fusion parameters, such as covariances or filter gains.
The bias-aware measurement model improves accuracy and estimator consistency relative to range-only formulations and remains effective with sparse anchors and a reduced tag set. Filter validation on an independent forklift dataset from a separate site further confirms reliable tracking across platforms.
In summary, the key contributions are:
\begin{itemize}
    \item Implementation of a novel deployment pipeline for UWB localization, combining online initialization and a generic multi-sensor fusion framework for localization.
    \item Analytical adaptation and practical integration of a bias-aware range measurement model into a \emph{manifold-based} estimator for AMRs.
    \item Real-world evaluation of the end-to-end pipeline on a commercial logistics AMR in a warehouse setting, assessing localization performance, model adaptations, and tag count reduction, with estimator-level cross-platform validation on a pre-calibrated second system.
    \item Publication of the warehouse trial data as a compact, ground-truthed resource for UWB–odometry fusion on an industrial AMR, available at: \textit{http://sst.aau.at/cns/datasets}
\end{itemize}

\section{Related Work}
\label{sec:related}
Foundational work on UWB-based localization typically separates calibration of tag extrinsics or measurement parameters from pose estimation, solving one while assuming the other~\cite{Ridolfi2021UWBSurvey}. Recent work increasingly targets end-to-end pipelines that combine both tasks to simplify deployment and improve positioning. Some methods retain a two-stage structure, with offline calibration followed by online localization, for example by combining UWB ranging with visual–inertial odometry (VIO)~\cite{Nguyen2021UWB_VIO_fusion_with_initialization} or LiDAR–IMU–UWB initialization with one-shot localization~\cite{Yuan2025UWB_IMU_Lidar_calib_loc}. Others pursue tightly coupled methods that estimate calibration and robot states simultaneously, such as the UWB-VIO fusion proposed in~\cite{Hu2024UWB_VIO_with_online_calib}. Nevertheless, calibration and localization are still often addressed by dedicated methods.

Utilizing ranges between static UWB anchors and onboard tags requires known anchor positions and pairwise two-way-ranging (TWR) biases. UVIO~\cite{Delama2023UVIO:Initialization} combines UWB ranging with VIO for low-drift localization and anchor calibration, using a least-squares formulation to estimate anchor positions and biases. The framework was later extended to support multiple tags and meshed bidirectional ranging between known and unknown anchors~\cite{Jung2024ModularCalibrationNew}. In this work, we use a custom unidirectional meshed TWR protocol to obtain range measurements from multiple tags to a time-varying set of anchors. 
Further refinements in~\cite{Delama2025Real-TimeNavIROS} improve calibration by considering positional dilution of precision (PDOP), lightweight measurement rejection, and an adaptive robust kernel for nonlinear optimization.

Localization can be performed with UWB alone through direct trilateration~\cite{Ruiz2017Commercial_UWB_analysis} or learning models trained on trilateration outputs~\cite{Poulose2020ML_UWB_Loc}. In practice, sensor fusion is often preferred because it mitigates limitations of individual modalities and improves accuracy and robustness. UWB ranges may be used loosely coupled, for example by feeding pre-filtered trilateration results to an EKF~\cite{Tran2025Loosly_coupled_UWB_fusion}. Tightly coupled integration directly processes raw ranges~\cite{Hol2009Tightly_coupled_UWB_fusion} and implicitly accounts for PDOP, which has been shown to be beneficial in industrial AMR scenarios~\cite{Dwek2020UWB_Loc_Industrial_UGV}.
Other fusion approaches include specialized filters, such as fuzzy adaptive filtering for indoor AMRs~\cite{Liu2019Fuzzy_UWB_IMU_fusion_UGV}, as well as particle filter (PF) and Kalman Filter (KF) variants. PFs have been applied to IMU–UWB person tracking~\cite{Tian2020PF_UWB_IMU_Person_tracking} and cooperative localization with optimization-based methods~\cite{Han202AntColonyPF_UWB_fusion}.

With the estimation problem being inherently nonlinear, KF variants remain widely used. EKFs are common for fusing UWB with IMU or odometry data in AMRs\cite{Dwek2020UWB_Loc_Industrial_UGV,Jang2024_UWB_IMU_Odometry_fusion_ekf} and iterated EKFs have been studied for forklift localization in simulated industrial environments~\cite{Barral2019IEKF_Simulated_Industrial_Forklift_with_UWB}. Sigma-point filters, including unscented and cubature KFs, have also been considered\cite{Liu2024UWB_odometry_fusion_ukf,Dong2023CubatureKF_UWB_fusion}.
Recent mobile-robot localization methods additionally exploit state geometry or environmental structure, such as stochastic filtering on the Lie group $\mathrm{SE}_2(3)$~\cite{Hashim2024Stochastic_UWB_IMU_fusion_on_Lie_Group} and manifold-based approaches such as PFs or Manifold Invariant EKFs for magnetic AMRs operating on metal structures~\cite{Chahine2022UWB_manifold_pf_imu_odometry,Starbuck2021UWB_MIEKF}. Following the same principle, a Manifold Error-State EKF (M-ESEKF) has been proposed to implicitly capture the surface-bound motion of AMRs~\cite{Raab2025MESEKF}.

In this paper, we extend the latter by introducing a bias-aware range measurement model based on~\cite{Blueml2021UWBBias} and by accounting for calibration uncertainties of UWB transceiver positions (see \ref{ssec:measurement_model}). These parameters are treated as auxiliary sensor states to enable seamless integration with an automatic anchor calibration stage, yielding a self-configuring, two-step pipeline featuring separate initialization and estimation. As shown in Sec. \ref{sec:results}, the measurement model modifications improve localization accuracy and estimator consistency, while auto-calibration of UWB parameters enables practical deployment with minimal user intervention.

\section{Notation}
\label{sec:notation}
In this paper bold lowercase letters represent vector quantities and bold capital ones denote matrices. Frames of reference are denoted as $(A)$ and $(B)$. Vectors representing physical quantities in a frame of reference $(A)$ are written as ${}_A\bm{r}$.
Rotation matrices that describe the orientation of a frame of reference $(B)$ relative to a frame $(A)$ are denoted by $\bm{R}_{AB}$.
Specifically, $\bm{R}_{AB}$ transforms a vector ${}_B\bm{r}$ expressed in $(B)$ into a vector ${}_A\bm{r} = \bm{R}_{AB} {}_B\bm{r}$ expressed in $(A)$.
\section{Methodology}
\label{sec:method}
Although the initialization and fusion strategies build on established methods, this section presents the necessary steps for their seamless integration into a unified pipeline. Particular attention is given to the adaptations and details that enable practical, largely automated deployment. 

\begin{figure}[]
    \centering
    \vspace{1em}
    \includegraphics[width=1\linewidth]{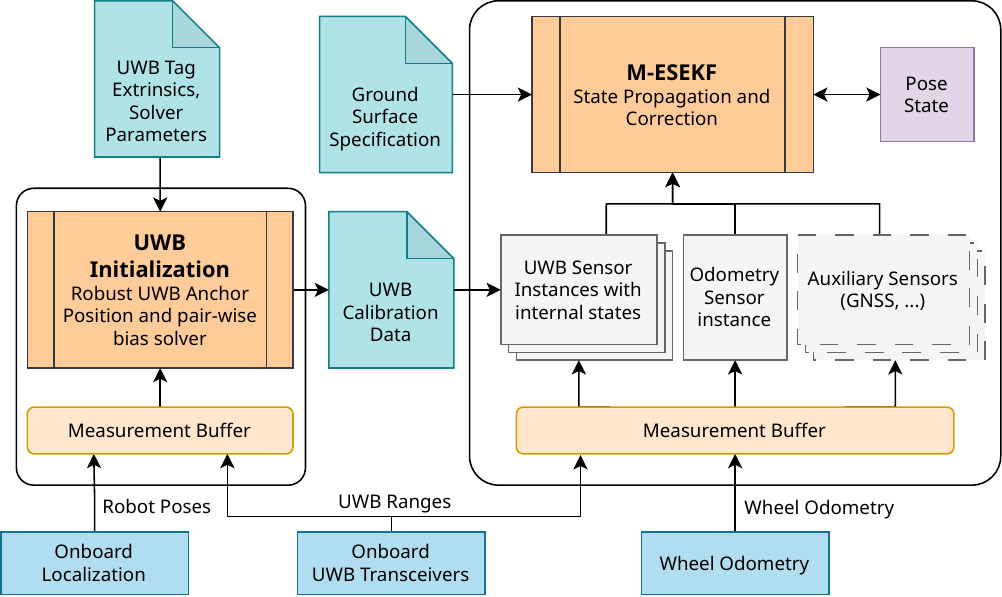}
    \vspace{-1.7em}
    \caption{Overview of the proposed pipeline combining single-step UWB calibration (left) with a modular, terrain-aware EKF (right) for pose tracking. First, anchor positions and pair-wise transceiver biases are estimated from UWB ranges and an auxiliary localization system supplying robot poses. The M-ESEKF then utilizes the calibration results for initializing sensor instances used for surface-aware pose estimation via UWB-odometry fusion.}
    \label{fig:pipeline_overview}
    \vspace{-1.5em}
\end{figure}

The proposed pipeline comprises two stages, shown in Fig.~\ref{fig:pipeline_overview}. The first performs automatic calibration of static UWB transceivers using AMR pose priors for instance provided by onboard localization systems, UWB ranges, and known tag extrinsics. 
It estimates anchor positions with covariances that capture unidirectional uncertainties and identifies pairwise constant and range-dependent biases, each with associated covariances. The resulting calibration file initializes the UWB sensor instances in the estimator for the second stage. Localization then employs a terrain-informed M-ESEKF that enforces surface-bound motion using a smooth ground model, such as B-spline approximations for UWB-odometry fusion. Both stages operate without extensive manual tuning and the following subsections provide details on the calibration and fusion procedure.
\subsection{Real-Time Initialization}
Leveraging meshed UWB ranging with runtime-deployed anchors in the proposed estimator requires estimates of anchor positions and pairwise biases.
These are obtained with the robust nonlinear least-squares solver from~\cite{Jung2024ModularCalibrationNew} using AMR pose estimates, known onboard UWB tag extrinsics, and ranges collected to newly deployed anchors along a short calibration trajectory. Therefore calibration quality depends on the accuracy of the pose priors and the excitation through the trajectory.
The estimated means and covariances are then passed to the sensor-fusion framework.
Unlike~\cite{Jung2024ModularCalibrationNew}, a different custom two-way-ranging firmware was used.
It builds upon a configurable Time-Division Multiple Access (TDMA) scheme, allowing each UWB tag to compute ranges in a certain interval to a set of identified UWB anchors.
The net TWR rate for a three-tag configuration is roughly 300 ranges per second in line-of-sight (LOS) conditions.
\subsection{Terrain-Informed Sensor Fusion Framework}
\label{ssec:mesekf}

\begin{figure}[b]
    \vspace*{-2em}
    \centering
    \includegraphics[width=.9\linewidth, trim={5cm 10.8cm 5cm 12cm},clip]{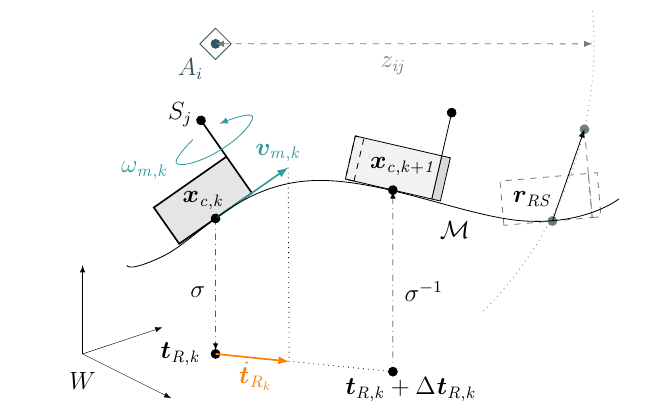}
    \vspace*{-3mm}
    \caption{The state space model on the manifold $\mathcal{M}$ and a range measurement $z_{ij}$ between anchor $A_i$ and onboard tag $S_j$ as utilized by the M-ESEKF. The robot state $\bm{x}_{c,k}$ at time step $k$ is propagated via a projection of the linear, and locally planar odometry velocity $\bm{v}_{m,k}$ onto the $xy$-plane via $\sigma$, forming $\dot{\bm{t}}_{R_k}$, while angular ones $\omega_{m,k}$ are processed in the tangent plane. This ensures viable states $\bm{x}_{c,k+1}$ in $\mathbb{R}^3$ via the inverse mapping $\sigma^{-1}$.}
    \label{fig:mesekf_definitions}
\end{figure}

To increase overall accuracy and robustness while enabling integration with generic robot models, the ranging data from the previously initialized UWB tags is fused with additional sensor modalities within an EKF.
Surface-bound AMR motion is handled inherently with an M-ESEKF~\cite{Raab2025MESEKF}, avoiding manually tuned artificial constraints.
The formulation assumes a known, sufficiently smooth surface model, which is typical for industrial environments with mostly flat floors and moderate ramps.
Such terrain can be represented analytically as a surface manifold $\mathcal{M}$.
Consequently, the nominal robot pose state $\bm{x}_c = (\bm{t}_R,\,\theta_R) \in \Rn{3}$ is fully described by its planar position $\bm{t}_R \in \Rn{2}$, heading angle $\theta_R \in \R$, and the explicit surface elevation model $z = S(\bm{t}_R) \in \R$.
The chart map $\sigma$ and its inverse $\sigma^{-1}$ provide a mapping between these domains as illustrated in Fig. \ref{fig:mesekf_definitions}.
In particular, a 3D point on the manifold $\bm{p}_R = (x,\,y,\,z) \in \mathcal{M}$ is mapped to its planar coordinates $\bm{t}_R$ with the chart map $\sigma : \mathcal{M} \to \Rn{2}$, defined as
\begin{gather}
    \sigma(\bm{p}_R) = \begin{bmatrix} 1 & 0 & 0 \\ 0 & 1 & 0 \end{bmatrix}\bm{p}_R = \bm{t}_R = \begin{bmatrix} x \\ y \end{bmatrix},
\end{gather}
The inverse mapping $\sigma^{-1} : \Rn{2} \to \mathcal{M}$ is defined as
\begin{gather}
    \sigma^{-1}(\bm{t}_R) = \bm{p}_R = \begin{bmatrix} x & y & S(x,y) \end{bmatrix}^\top ,
\end{gather}
where $S(x,y) = z$ is the surface elevation model.
The full orientation with respect to the world frame $(W)$ is decomposed into roll and pitch components from the ground model, $\bm{R}_{WT}$, and a heading rotation $\bm{R}_{TR}(\theta_R)$ about the local z-axis within the 
tangent plane:
\begin{gather}
    \bm{R}_{WR} = \left.\bm{R}_{WT} \right|_{\bm{p}} \bm{R}_{TR}(\theta_R).
\end{gather}
In addition to the state-space model, \cite{Raab2025MESEKF} outlines the widely adopted approach of modeling wheel odometry measurements as locally planar velocities for state propagation.
To improve long-term tracking, these models are integrated into an Error State EKF (ESEKF), as introduced by \cite{Roumeliotis1999ESEKF}, and implemented within the modular state estimation framework \textit{MaRS}.
The framework partitions the full system state into the robot pose $\bm{x}_c$, the core state, and sensor-specific states $\bm{x}_s$, such as calibration parameters. Each filter correction operates only on the relevant state and covariance sub-blocks, substantially reducing the effective dimensionality per update. Sensor–sensor cross-covariance blocks between different sensor instances are neglected. This enables consistent pose estimation across different platforms while efficiently accommodating the many sensor-specific states required for extensive UWB networks \cite{Brommer2021Mars}.

\subsection{Range Measurement Model}
\label{ssec:measurement_model}
The original M-ESEKF formulation \cite{Raab2025MESEKF} already considers pure range measurements between two devices as potential inputs for pose estimation
\begin{gather}
    z_{ij} = \|\sigma^{-1}(\bm{t}_R) + \bm{R}_{WR}{}_R\bm{r}_{RS,j}-{}_W\bm{r}_{A,i}\|_2.
\end{gather}
Here, ${}_W\bm{r}_{A,i}$ denotes the position of anchor tag $i$ in the world frame $(W)$, and ${}_R\bm{r}_{RS,j}$ is the offset of onboard tag $j$ in the robot frame $(R)$. 

This ideal range model is insufficient for real deployments because UWB transceiver positions are uncertain and two-way-ranging systems exhibit both constant bias $\beta$  and range-dependent bias $\gamma$~\cite{Blueml2021UWBBias}. The model is therefore extended with an augmented sensor state $\bm{x}_{s}$ that treats anchor positions, onboard tag offsets, and pairwise biases as probabilistic variables. The parametrization contains a set $\bm{r}_A$ of $m$ anchor positions ${}_W\bm{r}_{A,i}$, a set $\bm{r}_{RS}$ of $n$ onboard tag offsets ${}_R\bm{r}_{RS,j}$, and two sets of $mn$ scalar pairwise bias parameters, $\bm{\beta}, \bm{\gamma}$. The overall state $\bm{x}$ is formed by stacking these quantities
\begin{gather}
    \bm{r}_A = \begin{bmatrix}
        {}_W\bm{r}_{A,1}^\top & \dots & {}_W\bm{r}_{A,m}^\top
    \end{bmatrix}^\top \in \R^{3m}, \\
    \bm{r}_{RS} = \begin{bmatrix}
        {}_R\bm{r}_{RS,1}^\top & \dots & {}_R\bm{r}_{RS,n}^\top
    \end{bmatrix}^\top \in \R^{3n}, \\
    \bm{\beta} = \begin{bmatrix}
        \beta_{11} & \beta_{12} \dots & \beta_{21} & \dots & \beta_{mn}
    \end{bmatrix}^\top \in \R^{mn}, \\
    \bm{\gamma} = \begin{bmatrix}
        \gamma_{11} & \gamma_{12} \dots & \gamma_{21} & \dots & \gamma_{mn}
    \end{bmatrix}^\top \in \R^{mn}, \\
    \bm{x}_{s} = \begin{bmatrix}
        \bm{r}_A^\top & \bm{r}_{RS}^\top & \bm{\beta}^\top & \bm{\gamma}^\top
    \end{bmatrix}^\top, \label{eq:sensor_states}\\
    \bm{x} = \begin{bmatrix}
        \bm{x}_{c}^\top & \bm{x}_{s}^\top
    \end{bmatrix}^\top.
\end{gather}
To process an individual range measurement, the estimation framework assembles a partial sensor state $\bm{x}_{s,ij}$ for a specific anchor–tag pair $ij$, while assuming no cross-correlations between these auxiliary state blocks 
\begin{gather}
    \bm{x}_{s,ij} = \begin{bmatrix}{}_W\bm{r}_{A,i}^\top & {}_R\bm{r}_{RS,j}^\top & \beta_{ij} & \gamma_{ij}\end{bmatrix}^\top, \\
    z_{ij} = \gamma_{ij}\|\sigma^{-1}(\bm{t}_R) + \bm{R}_{WR}{}_R\bm{r}_{RS,j}-{}_W\bm{r}_{A,i}\|_2 + \beta_{ij}. \label{eq:z_dist_new}
\end{gather}
Note that biases are unique to each transceiver pair, as indicated by the indices.
The complete perturbed state for each filter update is thus given by 
\begin{gather}
    \hat{\bm{x}}_{ij} = \begin{bmatrix}\hat{\bm{x}}_c & \hat{\bm{x}}_{s,ij}\end{bmatrix}^\top, \\
    \delta\bm{x}_{ij}= \begin{bmatrix}\delta\bm{t}_R^\top & \delta\theta_R &\delta\bm{r}_{A,i}^\top, & \delta\bm{r}_{RS,j}^\top & \delta \beta_{ij} & \delta \gamma_{ij}\end{bmatrix}^\top, \\
    \delta\bm{x}_{ij} = \hat{\bm{x}}_{ij} - \bm{x}_{ij},
\end{gather}
where $\delta\bm{x}_{ij}$ denotes the error states and $\bm{x}_{ij}$ the nominal, noise-free ones \cite{Roumeliotis1999ESEKF}.
This formulation ensures compatibility with the calibration parameters provided by the initialization.

For integration into the M-ESEKF, the true measurement $\hat{z}_{ij}$ is obtained by injecting the perturbed state into \eqref{eq:z_dist_new} as
\begin{equation}
    \!\!\!\!\!\!\hat{z}_{ij}\!=\!\left(\gamma_{ij} + \delta \gamma_{ij}\right)
    \left\|  {}_{W}\hat{\bm{r}}_{R} + {}_{W}\hat{\bm{r}}_{RS,j} - {}_{W}\hat{\bm{r}}_{A,i} \right\|_2\!+ \beta_{ij} + \delta \beta_{ij},
    \label{eq:mesekf_range}
\end{equation}
where, to improve readability, terms expressing the perturbed robot position ${}_{W}\hat{\bm{r}}_{R}$, onboard tag offset ${}_{W}\hat{\bm{r}}_{RS,j}$, and anchor position ${}_{W}\hat{\bm{r}}_{A,i}$ in $(W)$ are stated explicitly as
\begin{align*}
    {}_{W}\hat{\bm{r}}_{R} &= \sigma^{-1}(\bm{t}_R + \delta\bm{t}_R), \\
    {}_{W}\hat{\bm{r}}_{RS,j} = \left.\bm{R}_{WT}\right|_{\bm{t}_R + \delta\bm{t}_R} &\bm{R}_{TR}(\theta_R + \delta\theta_R) \left({}_R\bm{r}_{RS,j} + \delta\bm{r}_{RS,j}\right), \\
    {}_{W}\hat{\bm{r}}_{A,i} &= {}_W\bm{r}_{A,i} + \delta\bm{r}_{A,i}.
\end{align*}
The nominal measurement $z_{ij}$ is recovered by setting all error states to zero, enabling the definition of the residual $\tilde{z}_{ij}$ and the measurement equation $h(\delta\bm{x}_{ij})$:
\begin{align}
    &z_{n,ij} = \left.z_{ij}\right|_{\delta\bm{x}_{ij} = 0},
    &&\tilde{z}_{ij} = h(\delta\bm{x}_{ij}) = \hat{z}_{ij} - z_{ij} .
\end{align}
The measurement Jacobian $\bm{H}_{ij}$ for the ESEKF is obtained via first-order Taylor expansion with respect to the error state $\delta\bm{x}_{ij}$. Since each range provides only a scalar constraint, some configuration parameters are weakly observable under generic AMR trajectories. Their uncertainties can therefore be managed more consistently using a Schmidt-Kalman approach instead of considering them as updated states \cite{Schmidt1966CEKF,Woodbury2010CEKF}. If not mentioned otherwise, the sensor states $\bm{x}_{s}$ in \eqref{eq:sensor_states} are treated as such Schmidt states, calibrated during initialization and held fixed in the measurement update while their cross-covariances with the core state are preserved. This avoids spurious updates of weakly observable parameters, but the capability to calibrate drifting parameters online is lost. Concretely, the uncertainty of these fixed Schmidt states is incorporated by inflating the measurement noise $\bm{R}_m$. Let $\bm{J}$ be the Jacobian of $\bm{h}$ with respect to the Schmidt parameters. Interpreting $\bm{J}$ as a first-order linear map into measurement space, their covariance $\bm{P}_c$ contributes an additional term, yielding the effective measurement covariance
\begin{gather}
    \bm{R} = \bm{R}_m + \bm{J}\bm{P}_c\bm{J}^\top.
\end{gather}
\vspace{-1.5em}

\section{Experiments and Results}
\label{sec:results}
This section evaluates the proposed approach on two mobile platforms. The first experiment demonstrates end-to-end deployment on a logistics AMR, while the second assesses estimator transferability using an independent, pre-calibrated forklift dataset. Performance is reported in terms of accuracy and consistency, with details provided in the following subsections.

\subsection{Omnidirectional Logistics AMR}
A commercial \textit{AGILOX ONE}\footnote{https://www.agilox.net/en/product/agilox-one/} AMR was used to evaluate the pipeline under representative warehouse conditions. This autonomous forklift operates on mostly flat industrial flooring, with permissible inclines typically below 1\%, and uses four independently steered drive units for quasi-omnidirectional motion. Its LiDAR-based onboard localization provides ground truth poses, enabling accurate and reproducible evaluation in realistic intralogistics scenarios.

\subsubsection{Setup}
The robot was equipped with three Qorvo DWM1001-DEV UWB transceivers (ID 1 to 3), mounted with an offset from the main chassis (Fig. \ref{fig:robots}) to reduce interference from metallic surfaces. The tags were arranged asymmetrically with large horizontal and vertical separations to improve PDOP and to support 3D pose estimation from ranges alone. Twelve UWB anchors were deployed throughout the warehouse, with devices ID 12 to 15 positioned outside the building to evaluate mixed indoor–outdoor behavior. Their ground-truth positions were aligned with geometric features extracted from the robot’s static, millimeter-precision map.
Fig. \ref{fig:agilox_trajectories} shows the environment, where black regions denote walls and gray regions indicate partial LOS obstructions caused by machinery and load carriers, particularly affecting the lowest tag. Within this setting, a dataset of 13 trajectories was recorded, containing 2D pose estimates from the LiDAR localization system, body velocities from the AMR, and unidirectional meshed range data from each onboard UWB tag to the anchors.

\subsubsection{Evaluation}
\begin{figure}[t]
    \centering
    \includegraphics[width=1\linewidth, trim={4.8cm 11.2cm 5cm 11.3cm},clip]{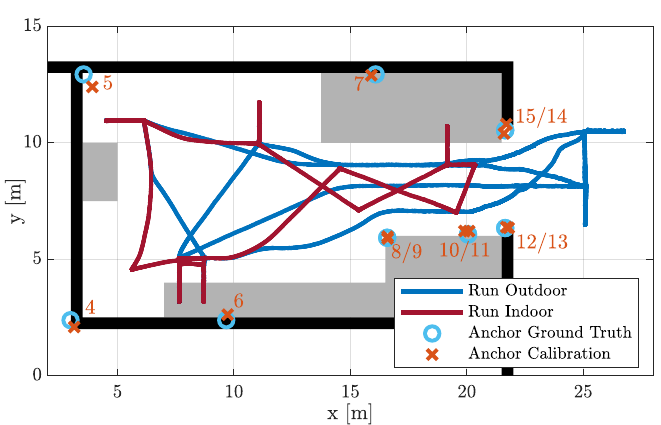}
    \vspace*{-8mm}
    \caption{The warehouse environment with two evaluation trajectories, anchor ground truths and their calibrated results. The indoor trial was used to record data for the anchor calibration, while the other run covers an extended area of operation to validate the estimation in previously unvisited space. Black areas visualize walls, while gray ones indicate partial LOS obstructions due to machinery and load carriers.}
    \label{fig:agilox_trajectories}
    \vspace{-1.5em}
\end{figure}

Two representative trajectories are analyzed in detail. One was confined to indoor operation and the other included outdoor sections, both shown in Fig. \ref{fig:agilox_trajectories}. The indoor trajectory corresponds to a typical load-carrier transport sequence and was used to calibrate anchors without specifically optimizing excitation or LOS conditions. The outdoor trial then evaluates whether accurate localization can be maintained in areas not used for initialization and with varying configurations.

Calibration quality is assessed against ground-truth anchor positions derived from the AMR's LiDAR map, while pose estimation is compared across measurement models and sensor configurations. Performance is quantified using the \textit{Absolute Trajectory Error} (ATE) of translation ($\mathrm{ATE}^\mathrm{t}$) and heading ($\mathrm{ATE}^\mathrm{\theta}$), as well as the \textit{Normalized Estimation Error Squared} (NEES)
\begin{gather}
    \bm{e}_{t, k} = \bm{t}_{R,k} - \bm{t}_{T,k} \in \R^2, \qquad  e_{\theta, k} = \theta_{R,k} - \theta_{T,k} \in \R^1,\\
    \bm{e}_{k} = \begin{bmatrix}  \bm{e}_{t, k}^\top & e_{\theta, k} \end{bmatrix}^\top \in \R^3, \\
    \mathrm{ATE}^\mathrm{t}_k = \left\| \bm{e}_{t, k}\right\|_2, \qquad \mathrm{ATE}^\mathrm{\theta}_k = \left| e_{\theta, k}\right|, \\
    NEES_k = \frac{1}{3}\bm{e}_{k}^\top\bm{P}_{k}^{-1}\bm{e}_{k}.
\end{gather}
Here, $\bm{e}_{t, k}$ and $e_{\theta, k}$ respectively denote the positional and orientational errors between the estimated states indexed $R$ and the ground truths signified by index $T$ at sample step $k$. ${\bm{e}_{k} \in \R^3}$ describes the combined pose error and ${\bm{P}_{k} \in \Rnm{3}{3}}$ the corresponding covariance. The ATE captures accuracy, whereas the NEES evaluates filter consistency. NEES values close to 1 indicate covariance estimates reflecting the true estimation error, thus better consistency~\cite{Li2012Credibility}.

\subsubsection{Results Indoor}
\begin{figure}[t]
    \centering
    \vspace*{2mm}
    \includegraphics[width=1\linewidth, trim={4.5cm 9.2cm 4.8cm 9.55cm},clip]{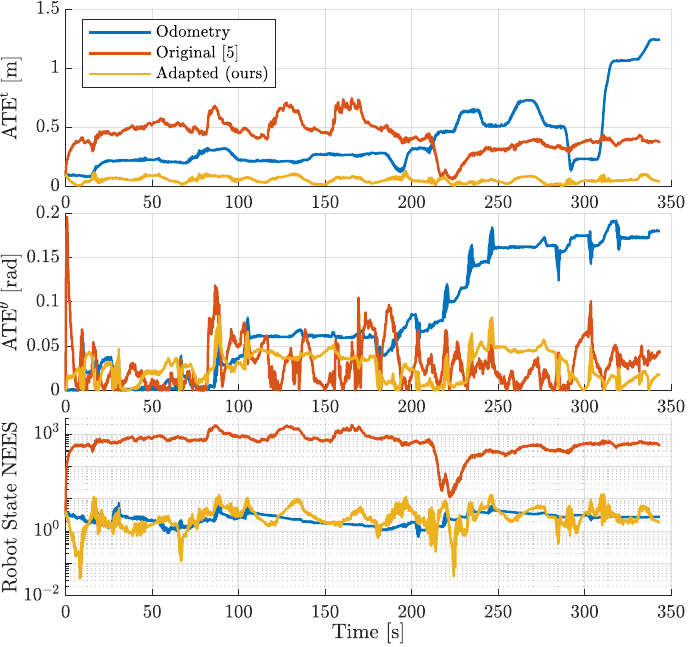}
        \vspace*{-8.5mm}
    \caption{Estimation results for odometry-only propagation, the original and the adapted range measurement formulation in the indoor setting. The plots indicate that the M-ESEKF is able to compensate the drift inherent to relative positioning with both measurement formulations, while our adaptation improves accuracy and estimator consistency.}
    \label{fig:agilox_indoor}
    \vspace{-1.4em}
\end{figure}

\begin{table}[b]
    \vspace{-1.4em}
    \centering
    \caption{Anchor Calibration Results on the indoor trajectory}
    \begin{tabular}{r|rrr|c|c}
         \multicolumn{1}{c}{} & \multicolumn{3}{c}{\textbf{Calibration Results}} & \multicolumn{1}{c}{\textbf{Abs. Error}} &  \multicolumn{1}{c}{\textbf{Est. Cov.}}  \\
        ID & $x$ [\si{m}] & $y$ [\si{m}] & $z$ [\si{m}] & [\si{m}] & [\si{m^2}] \\
        \midrule
        4  & 2.510 & 1.830 & 2.780 & 0.482 & 0.036 \\
        5  & 3.370 & 12.900 & 3.070 & 0.123 & 0.020 \\
        6  & 9.570 & 2.300 & 3.250 & 0.085 & 0.011 \\
        7  & 16.000 & 13.000 & 3.080 & 0.087 & 0.013 \\
        8  & 16.600 & 6.010 & 0.462 & 0.053 & 0.008 \\
        9  & 16.600 & 6.000 & 2.060 & 0.138 & 0.008 \\
        10 & 20.200 & 6.030 & 0.429 & 0.092 & 0.015 \\
        11 & 19.700 & 6.430 & 2.430 & \textcolor{gray}{0.525} & \textcolor{gray}{0.023} \\
        12 & 21.900 & 6.200 & 0.625 & 0.184 & 0.031 \\
        13 & 21.300 & 6.530 & 2.670 & 0.376 & 0.024 \\
        14 & 21.700 & 10.600 & 2.940 & 0.148 & 0.019 \\
        15 & 21.100 & 9.960 & 0.605 & \textcolor{gray}{0.467} & \textcolor{gray}{0.024} \\
        \bottomrule
        \multicolumn{6}{l}{\tiny{Grey values indicate, that the estimated position lies outside the $3\sigma$ bounds of the associated covariance.}}
    \end{tabular}
    \label{tab:agilox_calib}

\end{table}
Tab.~\ref{tab:agilox_calib} summarizes anchor calibration results, and Fig.~\ref{fig:agilox_indoor} compares odometry-only propagation, the original range formulation, and the adapted formulation that accounts for biases and calibration uncertainty through the Schmidt–Kalman approach. Overall, the calibration yields sufficiently accurate anchor positions, with most absolute position errors within the $3\sigma$-bounds of the estimated uniform covariances. Errors are larger in the z-coordinate than in the horizontal components, consistent with limited vertical excitation during flat-ground AMR operation. Raw UWB measurements show constant range biases up to $\SI{1.19}{\meter}$ between tags 3 and 15, while all range-dependent biases are calibrated to $1$ with covariances below $\SI{0.053}{}$.

Odometry-only filtering is slightly overconfident, with an average NEES of $2.5$. The original range model tracks the robot pose with maximum errors of $\SI{0.744}{\meter}$ and $\SI{11.23}{\degree}$, although estimator consistency deteriorates severely. In comparison, both accuracy and consistency improve substantially with the adapted formulation. Here, errors remain below $\SI{0.131}{\meter}$ and $\SI{4.81}{\degree}$, and mean NEES decreases from $625.3$ to $3.1$.

\subsubsection{Results Outdoor}
\begin{figure}[]
    \centering
    \vspace*{2mm}
    \includegraphics[width=1\linewidth, trim={4.5cm 9.3cm 4.8cm 9.55cm},clip]{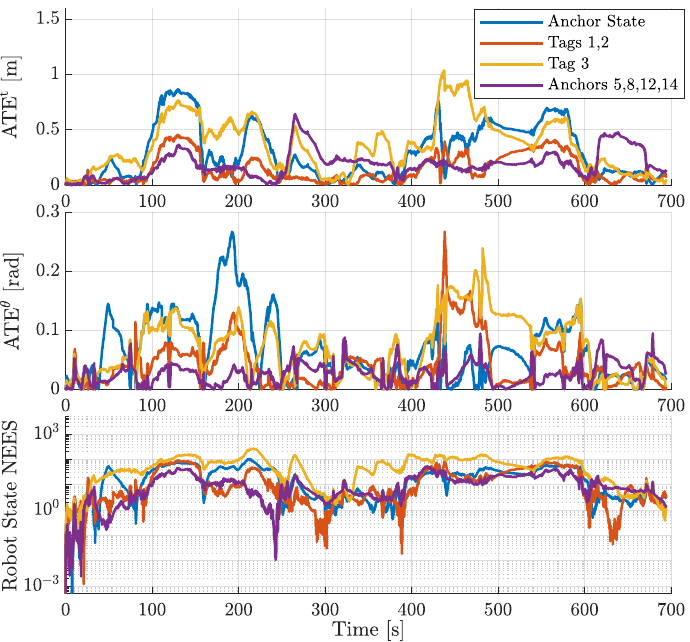}
        \vspace*{-8mm}
    \caption{AMR pose estimation results with anchor positions modeled as full states and using either all or a subset of active UWB devices with the adapted measurement formulation in the outdoor setting. Between $90-235\si{\second}$ and $405-600\si{\second}$ the robot is operating outside the building, possibly impacting odometry quality as the AMR is generally not intended for outdoor use.}
    \label{fig:agilox_outdoor}
    \vspace{-0.75em}
\end{figure}

The indoor calibration was reused for the outdoor trajectory to assess localization in an extended operation area, where minor floor unevenness introduces unmodeled disturbances under the flat-ground assumption. Fig.~\ref{fig:agilox_outdoor} shows representative results and Tab.~\ref{tab:agilox_outdoor_metrics} summarizes the corresponding metrics. Besides odometry-only, the original, and the adapted measurement formulations, the evaluation also included trials with adaptive anchor positions modeled as auxiliary sensor states, rather than fixed but probabilistic parameters. Additional experiments investigated the impact on performance when reducing the number of onboard tags or static anchors. During recording, anchor 13 malfunctioned and therefore did not provide data.

\begin{figure}[]
    \centering
    \vspace*{2mm}
    \includegraphics[width=1\linewidth, trim={4.2cm 12.8cm 4.8cm 12.61cm},clip]{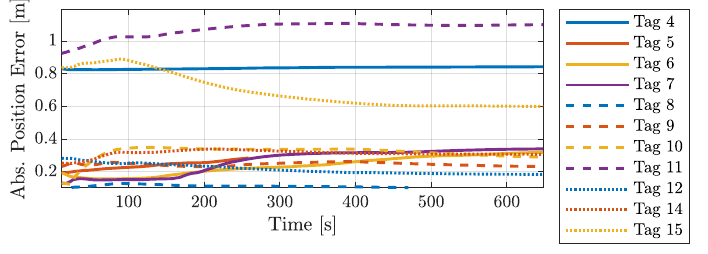}
        \vspace*{-8mm}
    \caption{Position error curves of the UWB anchors when treated as auxiliary sensor states on the outdoor trajectory, visualizing that while some initial calibrations can be improved online, most stay mostly unaffected or even worsen due to limited excitation of range-only measurements.}
    \label{fig:agilox_anchor_errors}
    \vspace{-1.4em}
\end{figure}

The results confirm that propagation itself remains credible on average, and the adapted measurement formulation improves performance compared to the original model, consistent with the indoor scenario. Absolute NEES values remain high however, indicating overconfidence likely caused by unmodeled non-LOS (NLOS) and multipath effects, surface-model mismatch, and neglected correlations between sensor-specific state blocks. Modeling anchor positions as auxiliary states improves consistency but reduces overall accuracy. The anchor error curves in Fig.~\ref{fig:agilox_anchor_errors} show that while some calibrations improve over time, most remain largely unchanged or degrade. The strongest changes occur for anchors 11 and 15, which are also the only anchors with calibrated positions outside the $3\sigma$-bounds of their estimated covariances. Modeling other parameters such as biases as updated states resulted in similar results. This likely stems from the observability conditions of the system. Biases remain weakly observable on generic AMR trajectories, since one-dimensional range measurements provide limited excitation and their effects are coupled with pose, tag, and anchor errors along the range direction. When anchor positions are modeled as updated states, the system becomes unobservable without fixed absolute references, since ranges constrain only relative geometry.

\begin{table}[]
    \centering
    \setlength{\tabcolsep}{5pt}
    \caption{Metrics of different configurations in the outdoor scenario}
    \begin{tabular}{lccc|lccc}
        \toprule
        Metric & Max & Mean & Std & Metric & Max & Mean & Std \\
        \midrule
        \multicolumn{4}{l}{\textbf{Odometry only}} & \multicolumn{4}{l}{\textbf{Original \cite{Raab2025MESEKF}}} \\ 
        $\mathrm{ATE}^\mathrm{t}$  & 2.869 & 0.637 & 0.748 & $\mathrm{ATE}^\mathrm{t}$  & 1.340 & 0.476 & 0.194 \\
        $\mathrm{ATE}^\mathrm{\theta}$  & 0.262 & 0.086 & 0.074 & $\mathrm{ATE}^\mathrm{\theta}$  & 0.422 & 0.080 & 0.085 \\
        NEES     & 5.668 & 0.951 & 0.548 & NEES     & 4342 & 858.3 & 634.0 \\
        \midrule
        \multicolumn{4}{l}{\textbf{Adapted (ours)}} & \multicolumn{4}{l}{\textbf{Anchor Position State}} \\
        $\mathrm{ATE}^\mathrm{t}$  & 0.555    & 0.175 & 0.148              & $\mathrm{ATE}^\mathrm{t}$             & 0.867 & 0.285 & 0.242 \\
        $\mathrm{ATE}^\mathrm{\theta}$  & 0.203    & 0.042 & 0.029       & $\mathrm{ATE}^\mathrm{\theta}$      & 0.267 & 0.065 & 0.054 \\
        NEES     & 192.7  & 39.48 & 47.02                               & NEES                               & 105.2 & 24.17 & 23.97 \\
        \midrule
        \multicolumn{4}{l}{\textbf{Anchors 5-8-12-14 only}} & \multicolumn{4}{l}{\textbf{Tag 1-2 only}} \\
        $\mathrm{ATE}^\mathrm{t}$  & 0.643 & 0.187 & 0.123     & $\mathrm{ATE}^\mathrm{t}$  & 0.456 & 0.136 & 0.123 \\
        $\mathrm{ATE}^\mathrm{\theta}$  & 0.095 & 0.027 & 0.018     & $\mathrm{ATE}^\mathrm{\theta}$  & 0.267 & 0.042 & 0.039 \\
        NEES     & 55.18 & 12.44 & 11.58  & NEES     & 93.17 & 19.12 & 23.82 \\
        \midrule
        \multicolumn{4}{l}{\textbf{Tag 1 only}} & \multicolumn{4}{l}{\textbf{Tag 3 only}} \\
        $\mathrm{ATE}^\mathrm{t}$  & 0.438 & 0.148 & 0.117 & $\mathrm{ATE}^\mathrm{t}$  & 1.037 & 0.355 & 0.239 \\
        $\mathrm{ATE}^\mathrm{\theta}$  & 0.252 & 0.044 & 0.037 & $\mathrm{ATE}^\mathrm{\theta}$  & 0.239 & 0.069 & 0.049 \\
        NEES     & 67.45 & 15.34 & 17.19 & NEES     & 269.1 & 64.7 & 55.26 \\
        \bottomrule 
        \multicolumn{8}{l}{\tiny{$\mathrm{ATE}^\mathrm{t}$ [\si{\meter}], $\mathrm{ATE}^\mathrm{\theta}$ [\si{\radian}]}}
    \end{tabular}
    \label{tab:agilox_outdoor_metrics}
    \vspace{-1.4em}
\end{table}

Reducing the number of onboard tags further enhances both accuracy and consistency. Since the M-ESEKF constrains the robot state to three degrees of freedom, two tags are theoretically sufficient to estimate the full 3D pose. Due to the off-center placement relative to the vehicle’s reference point, even a single tag can provide adequate localization performance. However, positioning based solely on tag 3 suffers from poor LOS and multipath effects due to its low mounting position, degrading accuracy. The observed improvements in consistency when using fewer tags may be explained by a reduction in conflicting information, as residual and position-dependent biases can distort the error distribution and lead to non-Gaussian measurement realizations.

Similarly, a considerably reduced number of active anchors can still support accurate positioning. The minimum number depends strongly on the environment, but in general, full coverage with at least two anchors in direct LOS of two onboard tags is required to fully observe the pose state on the ground-manifold. The trials using only anchors 5, 8, 12, and 14 illustrate this. Disabling further anchors severely degraded results and in some cases led to divergent estimates.

Prior works on anchor calibration with UAVs report mean anchor errors of $22-\SI{25}{\centi\meter}$ \cite{Delama2023UVIO:Initialization,Jung2024ModularCalibrationNew} and $11-\SI{21}{\centi\meter}$ with later refinements \cite{Delama2025Real-TimeNavIROS}. Despite reduced excitation on AMRs, our average error of $\SI{18}{\centi\meter}$ is within this range and improves over the $>\SI{40}{\centi\meter}$ reported for a short AMR trajectory in \cite{Delama2025Real-TimeNavIROS}. For UWB localization, our ATE of $18-\SI{36}{\centi\meter}$ and $2.4-\SI{4}{\degree}$ is slightly better than the VIO-aided UAV results in \cite{Delama2025Real-TimeNavIROS} with $34-\SI{57}{\centi\meter}$ and $3.6-\SI{4.1}{\degree}$. Fully comparable wheel-odometry and UWB fusion results in similarly industrial environments are however limited. Existing reports include median position errors below $\SI{6}{\centi\meter}$ in idealized lab setups \cite{Dwek2020UWB_Loc_Industrial_UGV} and a sub-$\SI{10}{\centi\meter}$ ATE in NLOS-heavy office settings \cite{Liu2024UWB_odometry_fusion_ukf}.

\subsubsection{Computational Effort}
To assess computational effort and real-time capability without a complete profiling study, we ran the filter offline on all recorded trajectories and measured compute time per step for the original and adapted formulations. On average propagation completes in under $\SI{3}{\micro\second}$ and UWB range updates finish in under $\SI{15}{\micro\second}$ with no notable difference between formulations. Most time is spent assembling and scattering per update state blocks, while the EKF computations take about $\SI{3}{\micro\second}$. This is consistent with prior reports on the efficiency of \textit{MaRS} when handling large sensor state sets \cite{Brommer2021Mars} and matches timings given in \cite{Raab2025MESEKF}.

\subsection{Experimental Forklift AMR}

\begin{figure}[]
    \centering
    \vspace*{2mm}
    \includegraphics[width=.85\linewidth, trim={4cm 11.1cm 4cm 12cm},clip]{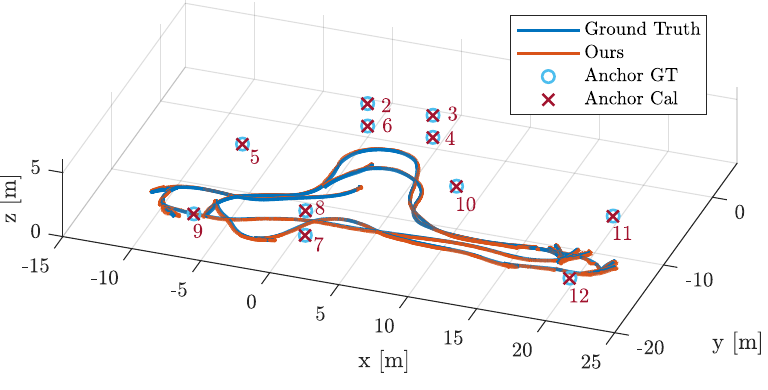}
    \vspace*{-4mm}
    \caption{The ground truth and the estimated trajectories for the cross-validation with the experimental forklift, including the true and calibrated anchor positions with IDs. The AMR has car-like steering and thus reverses multiple times to turn on the spot, which is noticeable close to $x = \SI{20}{m}$.}
    \label{fig:ait_trajectory}
    \vspace{-.75em}
\end{figure}

\begin{figure}[]
    \centering
    \includegraphics[width=1\linewidth, trim={4.2cm 10.4cm 4.8cm 10.5cm},clip]{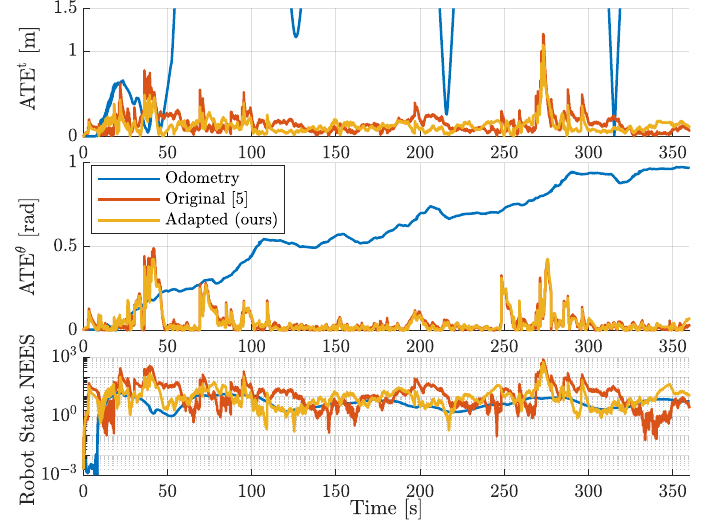}
    \vspace*{-8mm}
    \caption{Estimation results for odometry-only propagation, the original and the adapted range measurement formulation for the cross-validation. The odometry-only ATE grows outside the plot area and reaches values of up to $\SI{13.93}{\meter}$ and $\SI{56.32}{\deg}$ at $\SI{268.198}{\second}$. Accuracy and consistency of the original and the adapted measurement models are nearly equal, with the latter slightly outperforming in both metrics.}
    \label{fig:ait_results}
    \vspace{-1.4em}
\end{figure}

To assess transferability to different AMRs, the estimator was applied to a dataset recorded with an autonomous forklift \cite{Delama2025Real-TimeNavIROS} on near-flat ground. The platform provides RTK GPS–INS localization for ground truth and wheel odometry required by the M-ESEKF. UWB ranges to 12 static anchors were measured by two vertically spaced onboard tags, as illustrated in Fig.~\ref{fig:robots}. Recordings include pre-calibrated anchor positions with a maximum error of $\SI{0.05}{\meter}$. Constant UWB biases were derived offline from trajectory-range discrepancies. Range-dependent ones and terrain information were not available. Calibration results are shown in Fig.~\ref{fig:ait_trajectory} together with the reference and estimated trajectories. Fig.~\ref{fig:ait_results} reports the corresponding error and NEES curves, and Tab.~\ref{tab:ait_metrics} summarizes these metrics.

Odometry-only results exhibit substantial drift with ATEs of up to $\SI{13.93}{\meter}$ and $\SI{56.32}{\deg}$, suggesting less accurate relative positioning than in the logistics AMR. Both the original and the adapted M-ESEKF track the vehicle pose reliably, and the proposed modifications improve accuracy and NEES, although only modestly. This is likely due to higher-quality pre-calibration of anchor positions and constant biases, together with the lack of range-dependent bias estimation and larger unmodeled errors from the flat-ground approximation in this dataset. The latter two may also explain the overall higher NEES. Maximum ATE likely originates from odometry, as it occurs at the same time across all trials.

Although it does not re-evaluate the full deployment pipeline, the experiment indicates that the modified M-ESEKF generalizes well to different AMRs. Even with external pre-calibration and limited bias information, the adapted filter maintains reliable tracking and improves estimator performance compared to the original measurement models.

\begin{table}[t]
    \centering
    \vspace*{2mm}
    \setlength{\tabcolsep}{5pt}
    \caption{Evaluation metrics for Cross-Validation}
    \begin{tabular}{lccc|lccc}
        \toprule
        Metric & Max & Mean & Std & Metric & Max & Mean & Std \\
        \midrule
        \multicolumn{4}{l}{\textbf{Odometry only}} & \multicolumn{4}{l}{\textbf{Original \cite{Raab2025MESEKF} }} \\ 
        $\mathrm{ATE}^\mathrm{t}$  & 13.927 & 5.229 & 3.641 & $\mathrm{ATE}^\mathrm{t}$  & 1.228 & 0.150 & 0.132 \\
        $\mathrm{ATE}^\mathrm{\theta}$  & 0.983 & 0.584 & 0.285 & $\mathrm{ATE}^\mathrm{\theta}$  & 0.489 & 0.055 & 0.080 \\
        NEES   & 16.78 & 5.538 & 3.453 & NEES & 800.6 & 25.20 & 56.71 \\
        \midrule
        \multicolumn{4}{l}{\textbf{Adapted (ours)}} & \multicolumn{4}{l}{} \\
        $\mathrm{ATE}^\mathrm{t}$  & 1.061 & 0.1278 & 0.091  &&&&  \\
        $\mathrm{ATE}^\mathrm{\theta}$  & 0.426 & 0.0491 & 0.070  &&&& \\
        NEES     & 552.4 & 14.88& 30.95 &&&&  \\
        \bottomrule 
        \multicolumn{8}{l}{\tiny{$\mathrm{ATE}^\mathrm{t}$ [\si{\meter}], $\mathrm{ATE}^\mathrm{\theta}$ [\si{\radian}]}}
    \end{tabular}
    \label{tab:ait_metrics}
    \vspace{-1.4em}
\end{table}

\section{Conclusion}
\label{sec:conclusion}
This paper presents a two-stage UWB-aided localization pipeline for industrial AMRs with baseline localization and odometry. It combines automatic anchor configuration with terrain-informed UWB–odometry fusion through a bias-aware range model that incorporates calibration uncertainty. Experiments on a commercial logistics AMR show viable anchor initialization and accurate tracking indoors and across outdoor transitions, while reduced-anchor and reduced-tag trials demonstrate robustness under sparse UWB configurations. Cross-platform results on an independent forklift dataset further indicate estimator-level transferability.

The method assumes pose priors during initialization, a provided surface model, and platform-specific noise parameters. While limiting for generic applications, these are typically available in industrial deployments. Although the results demonstrate practical viability, further evaluation should include direct comparisons to external baselines such as tightly-coupled methods, as well as systematic studies regarding operating requirements like sensitivity to pose prior quality, surface-model mismatch and trajectory excitation. Future work could target extended automation through online surface and noise adaptation, the integration of additional sensors and improved handling of NLOS and multipath effects to enhance long-term robustness.

\section{Acknowledgements}
\textit{OpenAI ChatGPT 5.2} was used solely to improve grammar and spelling in the authors’ original text. All scientific content and interpretations are the authors’ own, and the authors take full responsibility for the final manuscript.

\bibliographystyle{IEEEtran}
\bibliography{bibliography/uwb.bib, bibliography/extra.bib}

\end{document}